\documentclass[10pt,twocolumn,letterpaper]{article}

\usepackage{cvpr}              
\usepackage{multirow}
\usepackage{booktabs}  
\usepackage{amsfonts} 
\usepackage{amsmath}
\usepackage{amssymb}
\usepackage{subcaption}
\usepackage{bm}
\usepackage{float}
\usepackage{graphicx}
\usepackage[font=small, skip=5pt]{caption}
\definecolor{cvprblue}{rgb}{0.21,0.49,0.74}
\usepackage[pagebackref,breaklinks,colorlinks,allcolors=cvprblue]{hyperref}

\def\paperID{7821} 
\def\confName{CVPR}
\def\confYear{2025}

\title{SemTalk: Holistic Co-speech Motion Generation with Frame-level Semantic Emphasis}

\author{
Xiangyue Zhang*$^{1,2}$ \quad Jianfang Li*$^2$ \quad Jiaxu Zhang$^{1}$ \\ \quad Ziqiang Dang$^{3,2}$ \quad Jianqiang Ren$^{2}$ \quad Liefeng Bo$^{2}$ \quad Zhigang Tu$^{1\dagger}$ \\
$^1$Wuhan University \quad $^2$Alibaba \quad $^3$Zhejiang University \\
\vspace{0.5em}
\textbf{Project page:} \href{https://xiangyuezhang.com/SemTalk/}{https://xiangyuezhang.com/SemTalk/}
}
\begin{document}

\twocolumn[{%
\renewcommand\twocolumn[1][]{#1}%
\maketitle
\centering
\begin{minipage}{0.95\textwidth}
\centering
\includegraphics[width=\textwidth]{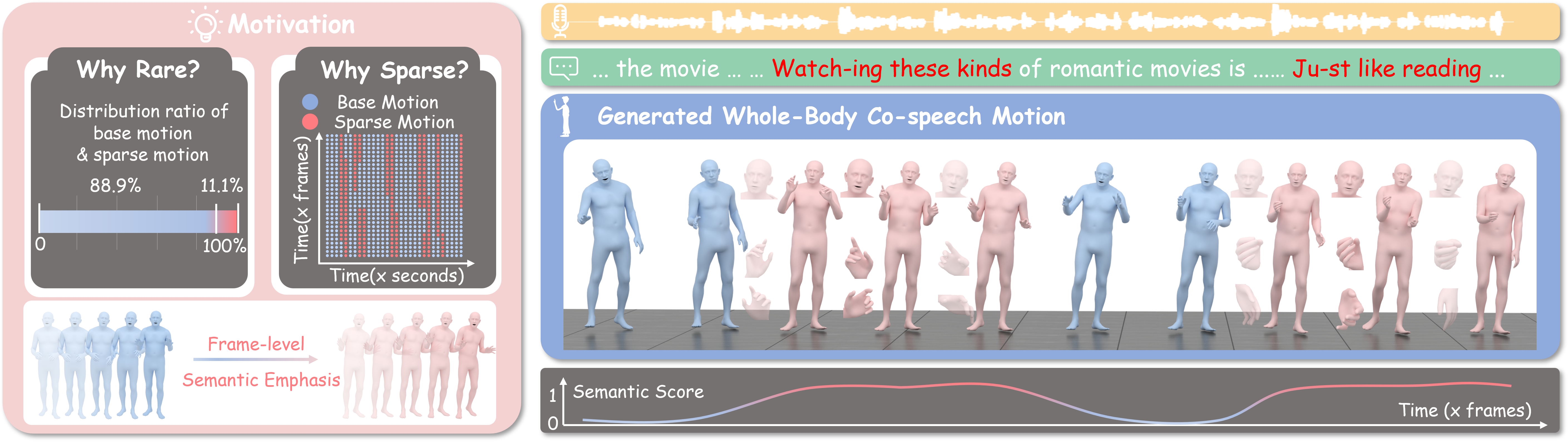}
\end{minipage}
\captionof{figure}{\textbf{On the left}, we analyze semantic labels from the BEAT2 dataset \cite{liu2024emage} and visualize frame-level motion, revealing that semantically relevant motions are rare and sparse, aligning with real-life observations. \textbf{On the right}, this observation drives the design of \textbf{SemTalk}, which establishes a rhythm-aligned base motion and dynamically emphasizes sparse semantic gestures at the frame-level. In this example, SemTalk amplifies expressiveness on words like “watching” and “just,” enhancing gesture and torso movements. The semantic scores below are automatically generated by SemTalk to modulate semantic emphasis over time. 
}
\label{fig:teaser}
}]

\renewcommand{\thefootnote}{\fnsymbol{footnote}} 
\footnotetext[1]{Xiangyue Zhang and Jianfang Li contributed equally to this work.}
\footnotetext[2]{Zhigang Tu is the corresponding author (\href{mailto:tuzhigang@whu.edu.cn}{tuzhigang@whu.edu.cn}).}
\renewcommand{\thefootnote}{\arabic{footnote}} 

\begin{abstract} Co-speech gesture generation must carefully integrate common rhythmic motion with rare yet essential semantic gestures. In this work, we propose SemTalk for holistic co-speech gesture generation with frame-level semantic emphasis. Our key insight is to separately learn base motions and sparse motions, and then adaptively fuse them. In particular, coarse2fine cross-attention module and rhythmic consistency learning are explored to establish rhythm-related base motion, ensuring a coherent foundation that synchronizes gestures with the speech rhythm. Subsequently, semantic emphasis learning is designed to generate semantic-aware sparse motion, focusing on frame-level semantic cues. Finally, to integrate sparse motion into the base motion and generate semantic-emphasized co-speech gestures, we further leverage a learned semantic score for adaptive synthesis. Qualitative and quantitative comparisons on two public datasets demonstrate that our method outperforms the state-of-the-art, delivering high-quality co-speech motion with enhanced semantic richness over a stable base motion. 
 \end{abstract}
    
\section{Introduction}
\label{sec:intro}
Nonverbal communication, including body language, hand gestures, and facial expressions, is integral to human interactions. It enriches conversations with contextual cues and enhances understanding among participants \cite{goldin1999role,kendon2004gesture,kucherenko2021large,cassell1999speech}. This aspect is particularly significant in holistic co-speech motion generation, where the challenge lies in synthesizing gestures that align with speech rhythm while also capturing the infrequent yet critical semantic gestures \cite{ozyurek2007line,lascarides2009formal}.

Most existing methods \cite{yi2023generating,liu2022beat,habibie2021learning} rely heavily on rhythm-related audio features as conditions for gesture generation. While these rhythm-based features successfully align gestures with the timing of speech, they often overshadow the sparse yet expressive semantic motion (see \cref{fig:teaser}). As a result, the generated motions may lack the contextual depth necessary and nuanced expressiveness for natural interaction.  Some methods try to address this by incorporating semantic information like emotion, style, and content\cite{chhatre2024emotional, danvevcek2023emotional, kucherenko2020gesticulator, liu2022learning}. However, the rhythm features tend to dominate, making the models difficult to capture sparse, semantically relevant gestures at the frame level. These rare but impactful gestures are often diluted or overlooked, highlighting the challenge of balancing rhythmic alignment with semantic expressiveness in co-speech motion generation.

In real-world human conversations, we have an observation that while most speech-related gestures are indeed rhythm-related, only a limited number of frames involve semantically emphasized gestures. This insight suggests that co-speech motions can be decomposed into two distinct components: (\(i\)) Rhythm-related base motion. These provide a continuous, coherent base motion aligned with the speech rhythm, reflecting the natural timing of speaking. (\(ii\)) Semantic-aware sparse motion: These occur infrequently but are essential for conveying specific meanings or emphasizing key points within the conversation.

Inspired by this observation, we propose a new framework SemTalk. SemTalk models the base motion and the sparse motion separately and then fuses them adaptively to generate high-fidelity co-speech motion. Specifically, we first focus on generating rhythm-related base motion by introducing \textit{coarse2fine cross-attention module} and \textit{rhythmic consistency learning}. We design a hierarchical \textit{coarse2fine cross-attention module}, which progressively refines the base motion cues in a coarse-to-fine manner, starting from the face and moving through the hands, upper body, and lower body. This approach ensures consistent rhythmic transmission across all body parts, enhancing coherence base motion.  Moreover, we propose a local-global \textit{rhythmic consistency learning} approach, which enforces alignment at both the frame and sequence levels. Locally, a frame-level consistency loss ensures that each frame is precisely synchronized with its corresponding speech features, guaranteeing accurate temporal alignment. Globally, a sequence-level consistency loss sustains a coherent rhythmic flow across the entire motion sequence, preserving consistency throughout the generated gestures.

Furthermore, we introduce \textit{semantic emphasis learning} approach, which focuses on generating semantic-aware sparse motion. This approach utilizes frame-level semantic cues from textual information, high-level speech features, and emotion to identify frames that require emphasis through a learned semantic score produced by a gating strategy, i.e., sem-gate. The sem-gate is designed to dynamically activate semantic motions at key frames through two weighting methods applied on the motion condition and the loss, respectively, and semantic label guidance, allowing the model to produce motion that enhances the motion with deeper semantic meaning and contextual relevance. 

Finally, the base motion and sparse motion are integrated through \textit{semantic score-based motion fusion}, which adaptively amplifies expressiveness by incorporating semantic-aware key frames into the rhythm-related base motion. 

Our contributions are summarized below:
\begin{itemize}
\item We propose SemTalk, a novel framework for holistic co-speech motion generation that separately models rhythm-related base motion and semantic-aware sparse motion, adaptively integrating them via a learned semantic gate.

\item We propose a hierarchical \textit{coarse2fine cross-attention module} to refine base motion and a local-global \textit{rhythmic consistency learning} to integrate latent face and hand features with rhythm-related priors, ensuring coherence and rhythmic consistency. We then propose \textit{semantic emphasis learning} to generate semantic gestures at certain frames, enhancing semantic-aware sparse motion.

\item Experimental results show that our model surpasses state-of-the-art methods qualitatively and quantitatively, achieving higher motion quality and richer semantics.
\end{itemize}

\section{Related Work}

\textbf{Co-speech Gesture Generation.} Co-speech gesture generation aims to produce gestures aligned with speech. Early rule-based methods \cite{kipp2005gesture,kopp2006towards,cassell2001beat,huang2012robot,shen2023difftalk} lacked variability, while deterministic models \cite{yoon2019robots,yang2023unifiedgesture,cassell1994animated,cassell2001beat,marsella2013virtual,liu2022disco} mapped speech directly to gestures. Probabilistic models, including GANs \cite{habibie2021learning,rebol2021real,ahuja2022low} and diffusion models \cite{zhu2023taming,alexanderson2023listen,yang2023diffusestylegesture,chhatre2024emotional}, introduced variability. Some methods incorporated semantic cues, such as HA2G \cite{liu2022learning} and SEEG \cite{liang2022seeg}, which used hierarchical networks and alignment techniques. SynTalker \cite{chen2024enabling} employs prompt-based control but treats inputs as signal strengths rather than fully interpreting semantics. LivelySpeaker \cite{zhi2023livelyspeaker} combines rhythmic features and semantic cues using CLIP \cite{radford2021learning} but struggles to integrate gestures with rhythm and capture semantics consistently, moreover, it only provides global control, limiting fine-grained refinement. DisCo \cite{liu2022disco} disentangles content and rhythm but lacks explicit modeling of sparse semantic gestures. SemTalk addresses this by separately modeling rhythm-related base motion and semantic-aware sparse motion, integrating them adaptively through a learned semantic score.

\noindent\textbf{Holistic Co-speech Motion Generation.} Generating synchronized, expressive full-body motion from speech remains challenging, especially in coordinating the face, hands, and torso \cite{lu2023humantomato,yi2023generating,liu2024emage,chen2024diffsheg,ng2024audio}. Early methods introduced generative models to improve synchronization, but issues persisted. TalkSHOW \cite{yi2023generating} improved with VQ-VAE \cite{van2017neural} cross-conditioning but handled facial expressions separately, causing fragmented outputs. DiffSHEG \cite{chen2024diffsheg} and EMAGE \cite{liu2024emage} used separate encoders for expressions and gestures, but their unidirectional flow limited coherence. ProbTalk \cite{liu2024towards} leverages PQ-VAE \cite{wu2019learning} for improved body-facial synchronization but mainly relies on rhythmic cues, risking the loss of nuanced semantic gestures. Inspired by TM2D \cite{gong2023tm2d}, which decomposes dance motion into music-related components, we separately model co-speech motion into rhythm-related and semantic-aware motion.
\section{Method}
\subsection{Preliminary on RVQ-VAE}
Following \cite{zeghidour2021soundstream,borsos2023audiolm,guo2024momask}, our approach uses a residual vector-quantized autoencoder (RVQ-VAE)  to progressively capture complex body movements in a few players. To retain unique motion characteristics across body regions, we segment the body into four parts—face, upper body, hands, and lower body—each with a dedicated RVQ-VAE, following \cite{ao2023gesturediffuclip,liu2024emage}. This segmentation preserves each part’s dynamics and prevents feature entanglement. 
\subsection{Overview}
\begin{figure}
    \centering
    \includegraphics[width=0.45\textwidth]{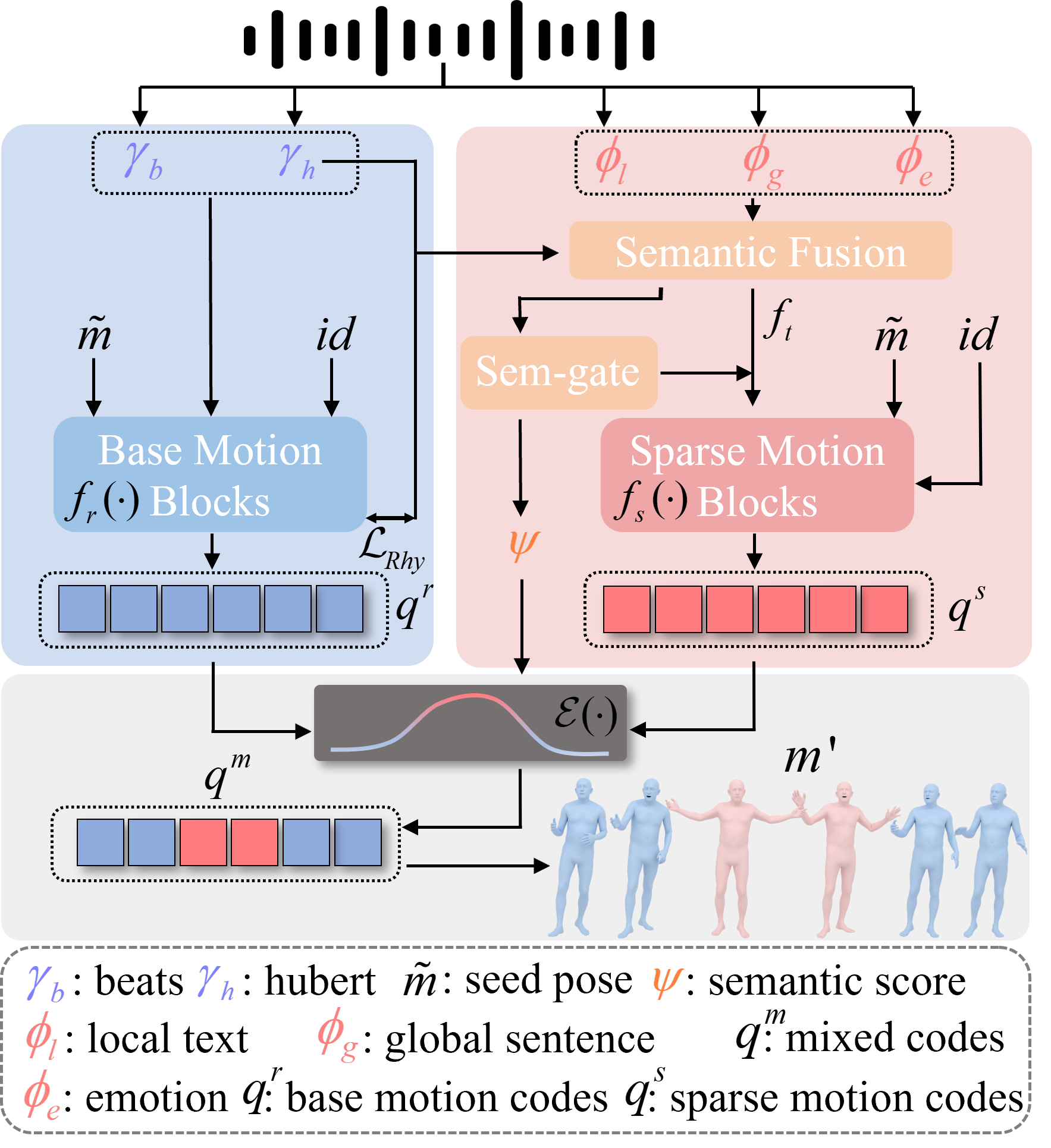}
    \caption{\textbf{An overview of the SemTalk pipeline.} SemTalk generates holistic co-speech motion by first constructing rhythm-aligned  $q^{r}$ in \( f_r \), guided by rhythmic consistency loss \( L_{\text{Rhy}} \). Meanwhile, \( f_s \) produce frame-level semantic codes \( q^s \), activated selectively by the semantic score \( \psi \). Finally, \( q^m \) is achieved by fusing $q^{r}$ and \( q^s \) based on \( \psi \), with motion decoder, yielding synchronized and contextually enriched motions.}
    \vspace{-15pt}
    \label{fig:SemTalk_overview}
\end{figure}

As shown in Figure \ref{fig:SemTalk_overview}, our SemTalk pipeline includes two main components: the Base Motion Blocks \( f_r(\cdot) \) and the Sparse Motion Blocks \( f_b(\cdot) \). Given rhythmic features \( \gamma_b \), \( \gamma_h \), a seed pose \( \tilde{m} \), and a speaker ID \( id \), the Base Motion Blocks generate rhythm-aligned codes \( q^b \), forming the rhythmic foundation of the base motion:
\begin{equation}
f_r : (\gamma_b, \gamma_h, \tilde{m}, id; \theta_{f_r}) \rightarrow q^b,
\end{equation}
where \( \theta_{f_r} \) denotes the learnable parameters of the Base Motion Blocks. The Sparse Motion Blocks then take semantic features \( \phi_l \), \( \phi_g \), \( \phi_e \), along with \( \gamma_h \), \( \tilde{m} \) and \( id \), to produce frame-level semantic codes \( q^s \) and semantic score \( \psi \). \( \psi \) then triggers these codes only for semantically significant frames, producing a sparse motion representation:
\begin{equation}
f_s : (\phi_l, \phi_g, \phi_e, \tilde{m}, id; \theta_{f_s}) \rightarrow (q^s, \psi),
\end{equation}
where \( \theta_{f_s} \) represents the Sparse Motion Block parameters. Finally, the semantic emphasis mechanism \(\mathcal{E}\) combines \( q^b \) and \( q^s \), guided by \( \psi \), to form the final motion codes \( q^m \):
\begin{equation}
q^m = \mathcal{E}(q^b, q^s; \psi).
\end{equation}
The motion decoder then uses \( q^m \) to generate the output \( m' \), enriched with rhythm alignment and semantic emphasis. 
\begin{figure*}
    \centering
    \includegraphics[width=0.9\textwidth]{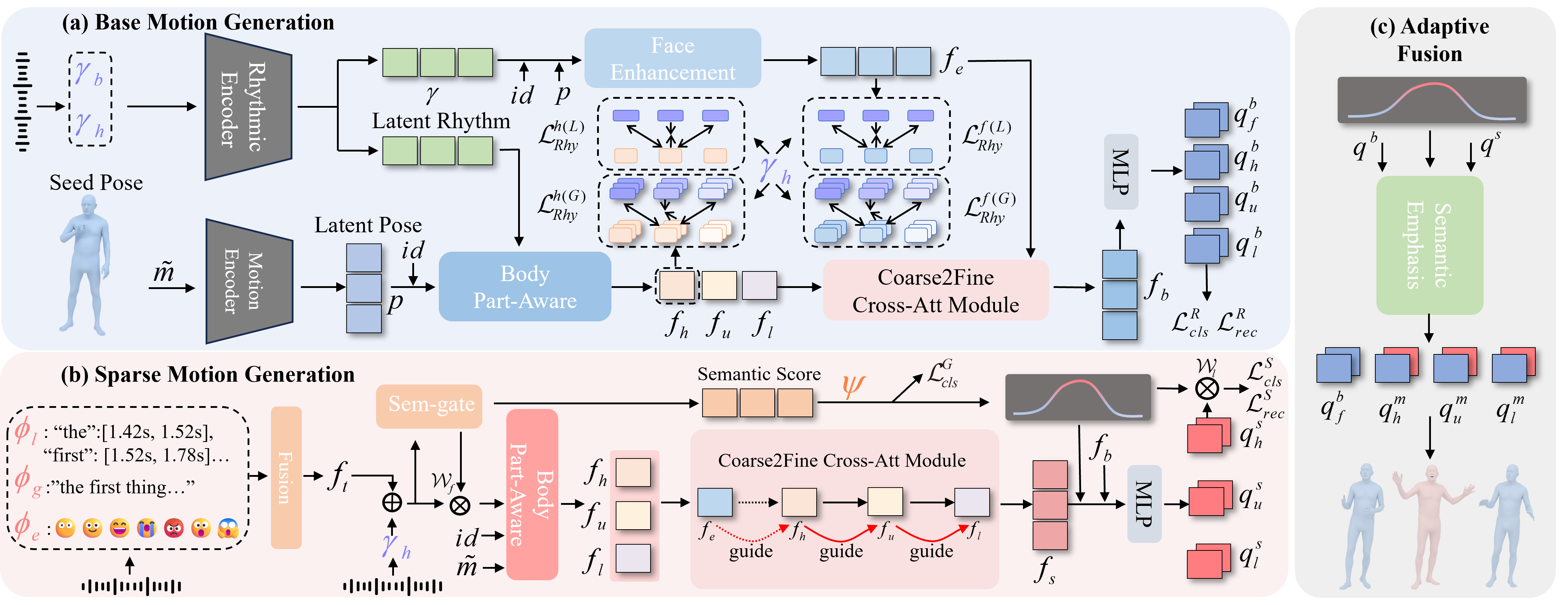}
    \caption{\textbf{Architecture of SemTalk.} SemTalk generates holistic co-speech motion in three stages. (a) Base Motion Generation uses \textit{rhythmic consistency learning} to produce rhythm-aligned codes \( q^b \), conditioned on rhythmic features \( \gamma_b \), \( \gamma_h \). (b) Sparse Motion Generation employs \textit{semantic emphasis learning}  to generate semantic codes \( q^s \), activated by semantic score \( \psi \). (c) Adaptively Fusion automatically combines \( q^b \) and \( q^s \) based on \( \psi \) to produce mixed codes \( q^m \) at frame level for rhythmically aligned and contextually rich motions.
 }
    \vspace{-15pt}
    \label{fig:method}
\end{figure*}

\subsection{Generating Rhythm-related Base Motion}
\label{sec:base_motion}
The Base Motion Generation (\cref{fig:method} a) in SemTalk establishes a rhythmically aligned foundation by leveraging both rhythmic and speaker-specific features, enhancing the naturalness and personalization of generated motion.

\noindent\textbf{Rhythmic Speech Encoding.} To synchronize motion with speech, SemTalk incorporates rhythmic features: beats \( \gamma_b \) and HuBERT features \( \gamma_h \). \( \gamma_b \), derived from amplitude, short-time energy \cite{chu2009environmental}, and onset detection, mark key rhythmic points for aligning gestures with speech. Meanwhile, \( \gamma_h \), extracted by the HuBERT encoder \cite{hsu2021hubert}, captures high-level audio traits. In addition to rhythmic features \(\gamma\), SemTalk uses a seed pose \( \tilde{m} \) and speaker identity \( id \) to generate a personalized, rhythm-aligned latent pose \(p\). Then MLP-based Face Enhancement and Body Part-Aware modules utilize \(\gamma\), \(p\) and \( id \) to obtain latent face \( f_e \), hands \( f_h \), upper body \( f_u \) and lower body \( f_l \). 

\noindent\textbf{Coarse2Fine Cross-Attention Module.} 
To facilitate the learning of base motion, we first proposed a transformer-based hierarchical \textit{Coarse2Fine Cross-Attn Module} utilize  \( f_e \),  \( f_h \),  \( f_u \) and \( f_l \) to obtain latent base motion \( f_b \). The refinement begins with \(\gamma\) for \( f_e \), which guides the rhythmic representation for \( f_h \), followed by conditioning \( f_u \) and finally influencing \( f_l \). Since mouth movements closely correspond to speech syllables with minimal delay, we use the face to guide hand motions, inspired by DiffSHEG \cite{chen2024diffsheg}. As the upper and lower body movements are less directly driven by speech and instead reflect the natural swinging of the hands and torso, we adopt cascading guidance: hands influence the upper body, which in turn drives the lower body.  This structured approach, moving from the face to the hands, upper body, and lower body, ensures smooth and coherent motion propagation across the entire body.

\noindent\textbf{Rhythmic Consistency Learning.} 
Inspired by CoG's use of InfoNCE loss \cite{xu2024chain} to synchronize facial expressions with audio cues, our approach adopts a similar philosophy of aligning motion and speech rhythm. It can be defined as:
\begin{equation}
\mathcal{L}_{\text{Rhy}} = -\frac{1}{N} \sum_{i=1}^N \log \frac{\exp \left(\operatorname{sim}\left(h\left(f_i\right), \gamma_h^{i}\right) / \tau \right)}{\sum_{j=1}^N \exp \left(\operatorname{sim}\left(h\left(f_i\right), \gamma_h^{j}\right) / \tau \right)},
\end{equation}
where \( N \) denotes the number of frames(or the batch size), \(\tau\) denotes the
temperature hyperparameter, \( h(\cdot) \) is the projection head for latent motion, \( f_i \) and \( \gamma_h^{i} \) are the latent motion and rhythmic features at frame (or sample) \( i \), and \( \operatorname{sim}(\cdot) \) represents cosine similarity.

Unlike CoG, our approach fundamentally differs by incorporating separate local and global rhythmic consistency losses, which are applied to both latent face \( f_e \) and latent hands  \( f_h \), ensuring a more cohesive and synchronized representation across the entire motion sequence. This rhythmic consistency loss ensures that the motions are not only synchronized at the frame level but also maintain a consistent rhythmic flow across the entire sequence. 

The local frame-level consistency loss \( \mathcal{L}_{\text{Rhy}}^{(L)} \) aligns the motion features of each frame with the corresponding rhythmic cues \(\gamma_{h}\). By leveraging HuBERT features \( \gamma_h \) instead of basic beat features \( \gamma_b \), which only capture rhythmic pauses, we incorporate rich, high-level audio representations that enhance the model’s ability to capture rhythm-related motion patterns and maintain temporal coherence. 

The global sentence-level consistency loss \( \mathcal{L}_{\text{Rhy}}^{(G)} \) is designed to ensure rhythmic coherence at a global level. Unlike local loss, \( \mathcal{L}_{\text{Rhy}}^{(G)} \) reinforces rhythm consistency throughout the sequence, ensuring that the generated motion maintains smooth and rhythm-aligned throughout its duration. 

By jointly minimizing \( \mathcal{L}_{\text{Rhy}}^{(L)} \) and \( \mathcal{L}_{\text{Rhy}}^{(G)} \), \textit{rhythmic consistency learning} enables SemTalk to produce base motions that are rhythmically aligned and temporally cohesive, forming a solid rhythm-related base motion foundation.

\subsection{Generating Semantic-aware Sparse Motion}  
The Sparse Motion Generation (\cref{fig:method} b) in SemTalk adds semantic-aware sparse motion to base motion by incorporating semantic cues drawn from speech content and emotional tone. By separating rhythm and semantics, this stage enhances motion generation by emphasizing contextually meaningful motion at key semantic moments.

\noindent\textbf{Semantic Speech Encoding.} To capture semantic cues in speech, \textit{Semantic Emphasis Learning} combines three types of information: frame-level text embeddings \( \phi_l \), sentence-level features \( \phi_g \) from the CLIP model \cite{radford2021learning}, and emotion features \( \phi_e \) from the emotion2vec model \cite{ma2023emotion2vec}. These features form a comprehensive semantic representation \( f_t \), together with audio feature \(\gamma_h\),  that reflects both the content and emotional undertones of speech, enabling SemTalk to activate motions that are sensitive to nuanced semantic cues.

\noindent\textbf{Semantic Emphasis Learning.} \label{sec:semantic-emphasis} The process begins by generating \( f_t \), combining local and global cues from text, speech, emotion embeddings and HuBERT features \(\gamma_{h}\). Then, the sem-gate leverages multi-modal inputs to generate a semantic score, identifying frames that require enhanced semantic emphasis. The sem-gate in SemTalk refines keyframe motion by applying two forms of weighting methods \(\mathcal{W}\): feature weighting \(\mathcal{W}_{f}\) and loss weighting \(\mathcal{W}_{l}\). Using \( f_t \) and \(\gamma_{h}\), SemTalk computes a semantic score \( \psi \), which dynamically scales feature weighting—filtering back semantic features \( f_t \) to activate frames with significant relevance, ensuring that the model emphasizes frames aligned with specific communicative intentions.  Second, the loss weighting is applied by supervising \( \psi \), with a classification loss \( \mathcal{L}_{cls}^{G}\) based on semantic labels, further enhancing the model’s ability to identify key frames. The two weighting methods allow SemTalk to selectively enhance semantic gestures while suppressing uninformative motion, leading to more expressive co-speech motion. 

Once \( \psi \) is established, it modulates the integration of rhythm-aligned base motion \( f_b \) and sparse semantic motion \( f_s \). Through alpha-blending, frames with high semantic relevance draw more from \( f_s \), while others rely on \( f_b \). The final motion codes \( q^s \) are computed as:

\begin{equation}
    q^s = MLP(\psi f_s + (1 - \psi) f_b),
\end{equation}

To ensure cohesive propagation of semantic emphasis across body regions, we employ the \textit{Coarse2Fine Cross-Attention Module}, similar to \cref{sec:base_motion}. In this stage, we focuses solely on body motion, excluding facial movements, as body gestures play a more critical role in conveying semantic meaning in co-speech interactions.

To foster diverse motion generation, SemTalk includes a code classification loss \( \mathcal{L}_{cls} \) and a reconstruction loss \( \mathcal{L}_{rec} \). These losses are specifically focused on frames with high semantic scores, guiding the model to prioritize the generation of sparse, meaningful gestures.
\begin{figure}
    \centering
    \includegraphics[width=0.45\textwidth]{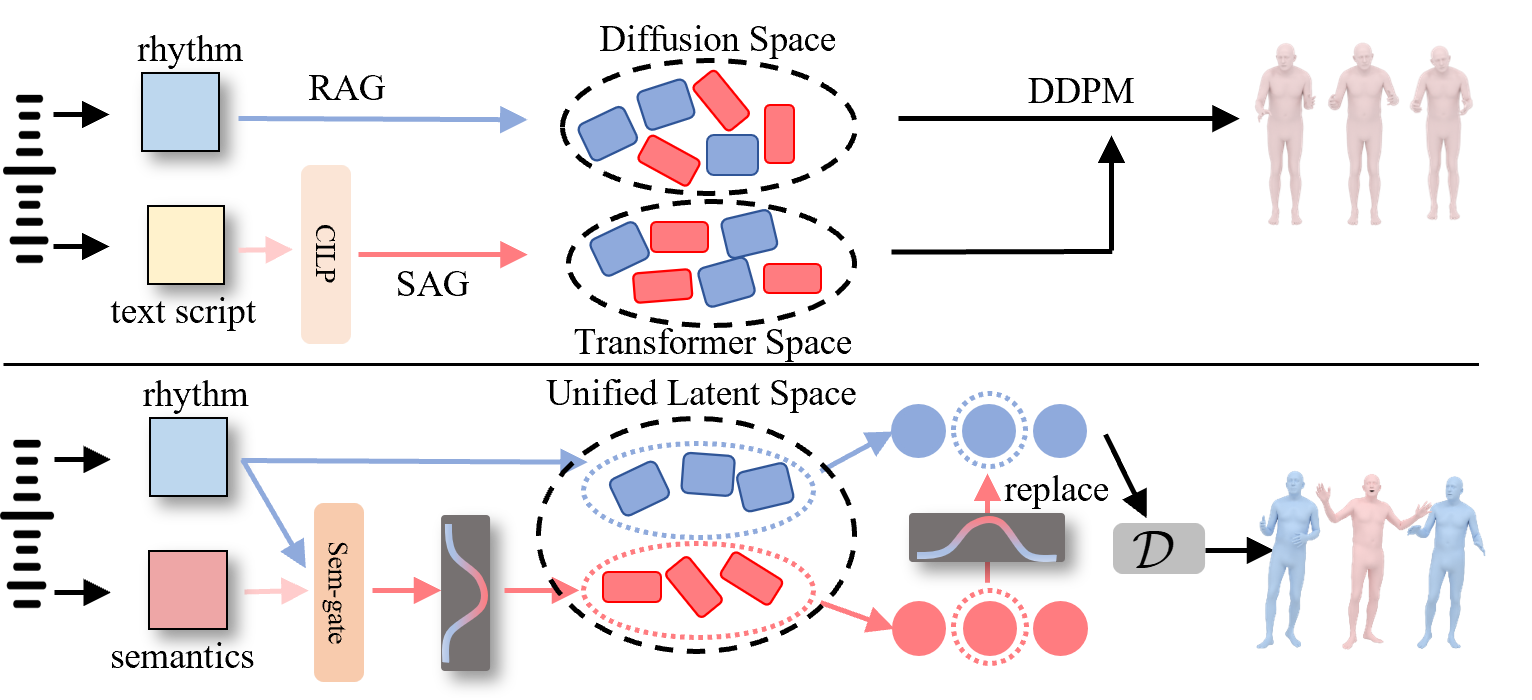}
    \caption{\textbf{Concept comparison with LivelySpeaker \cite{zhi2023livelyspeaker}.} (Top) LivelySpeaker generates semantic gestures with CLIP embeddings in SAG and refines rhythm-related gestures separately using diffusion, causing potential jitter. (Bottom) SemTalk integrates text and speech, uses a semantic gate for fine-grained control, and unifies rhythm and semantics for smoother, more coherent motions.}
    \vspace{-15pt}
    \label{fig:com_lively}
\end{figure}

\noindent\textbf{Discussion.} Recently, LivelySpeaker \cite{zhi2023livelyspeaker} designs the Semantic-Aware Generator (SAG) and Rhythm-Aware Generator (RAG) for co-speech gesture generation, combining them through beat empowerment. While effective, key differences exist between LivelySpeaker and SemTalk, see \cref{fig:com_lively}. First, SAG generates gestures from text using CLIP embeddings, but bridging words and expressive gestures is challenging, causing jitter. SemTalk incorporates speech features (pitch, tone, emotion) alongside text and GT supervision for adaptive gestures. Second, LivelySpeaker applies global control, missing local semantic details, while SemTalk uses fine-grained, frame-level semantic control for subtle variations. Third, LivelySpeaker fuses SAG and RAG in separate latent spaces, leading to misalignment and inconsistencies. SemTalk jointly models rhythm and semantics in a unified framework, ensuring smoother transitions and coherence. We further compare SAG with our semantic gate in experiments.

\begin{figure*}
    \centering
    \includegraphics[width=1\textwidth]{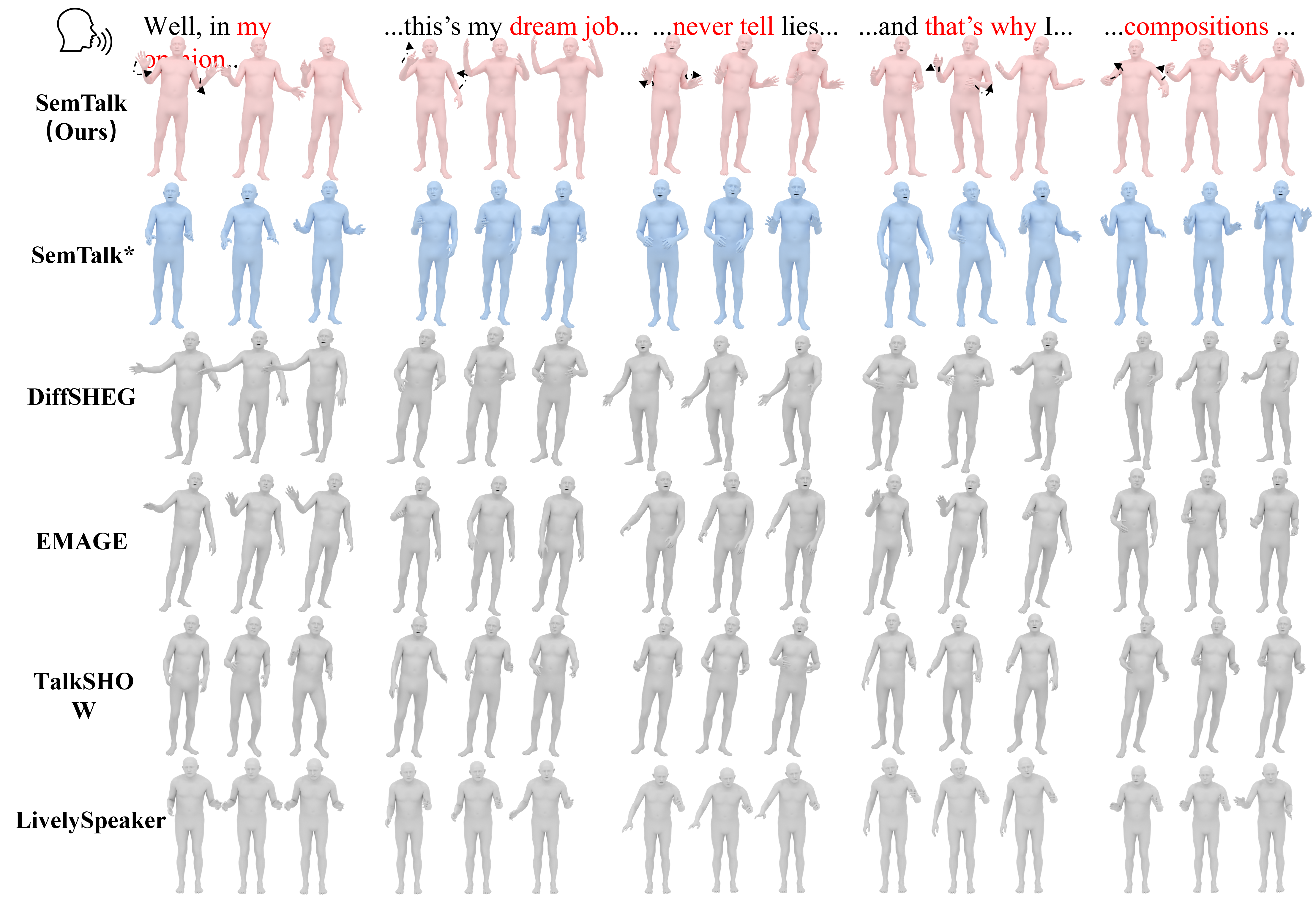}
    \caption{\textbf{Comparison on BEAT2 \cite{liu2024emage} Dataset.} SemTalk* refers to the model trained solely on the Base Motion Generation stage, capturing rhythmic alignment but lacking semantic gestures.  In contrast, SemTalk successfully emphasized sparse yet vivid motions. For instance, when saying “my opinion,” SemTalk generates a hand-raising gesture followed by an index finger extension for emphasis. Similarly, for “never tell,” our model produces a clear, repeated gesture matching the rhythm, reinforcing the intended emphasis.
}
\vspace{-10pt}
    \label{fig:compare}
\end{figure*}
\begin{figure*}
    \centering
    \includegraphics[width=0.9\textwidth]{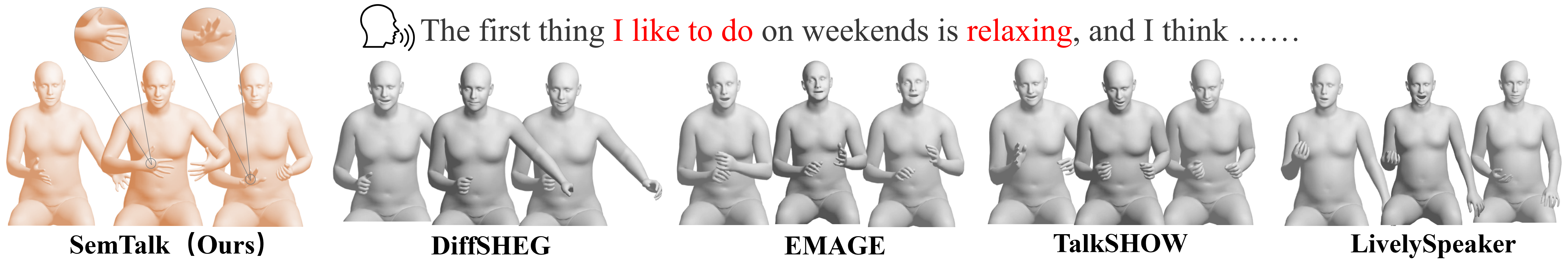}
    \caption{ \textbf{Comparison on SHOW \cite{yi2023generating} Dataset.} Our method performs better in motion diversity and semantic richness. 
}
\vspace{-20pt}
\label{fig:show_compare}
\end{figure*}
\subsection{Semantic Score-based motion fusion}

The Adaptive Fusion stage (\cref{fig:method} c) in SemTalk seamlessly integrates semantic-aware sparse motion into the rhythmic-related base motion. By strategically enhancing frames based on their semantic importance, it maintains a smooth and natural motion flow across sequences. For each frame \( i \), the semantic score \( \psi_i \) computed during the Sparse Motion Generation stage is compared to a threshold \( \beta \). If \( \psi_i > \beta \), the base motion's latent code \( q_i^r \) is replaced with the sparse semantic code \( q_i^s \), effectively highlighting expressive gestures where they are most relevant; otherwise, \( q_i = q_i^r \).

This selective replacement emphasizes semantically critical gestures while preserving the natural rhythmic base motion. By blending \( q^b \) and \( q^s \) based on semantic scores, SemTalk adapts to the expressive needs of the speech context while ensuring coherence. Additionally, the convolution structure of the RVQ-VAE decoder ensures smooth transitions between frames, preserving motion continuity. 
\section{Experiments}
\subsection{Experimental Setup}
\noindent \textbf{Datasets.} For training and evaluation, we use two datasets: BEAT2 and SHOW. \textbf{BEAT2}, introduced in EMAGE \cite{liu2024emage}, extends BEAT \cite{liu2022beat} with 76 hours of data from 30 speakers, standardized into a mesh representation with paired audio, text, and frame-level semantic labels. We follow \cite{liu2024emage} and use the BEAT2-standard subset with an 85\%/7.5\%/7.5\% train/val/test split. \textbf{SHOW} \cite{yi2023generating} includes 26.9 hours of high-quality talk show videos with 3D body meshes at 30fps. Since it lacks frame-level semantic labels, we use the sem-gate from SemTalk, pre-trained on BEAT2, to generate them. Following \cite{yi2023generating}, we select video clips longer than 10 seconds and split the data 80\%/10\%/10\% for train/val/test.

\noindent \textbf{Implementation Details.} Our model is trained on a single NVIDIA A100 GPU for 200 epochs with a batch size of 64. We use RVQ-VAE \cite{van2017neural}, downscaling by 4. The residual quantization has 6 layers, a codebook size of 256 and a dropout rate of 0.2. We use five transformer layers to predict the last five layer codes. In \textit{Base Motion Learning}, \(\tau\) = 0.1; in \textit{Sparse Motion Learning}, \(\beta\) = 0.5 empirically. The training uses ADAM with a 1e-4 learning rate. Following \cite{liu2024emage}, we start with a 4-frame seed pose, gradually increasing masked frames from 0 to 40\% over 120 epochs.

\noindent \textbf{Metrics.}We evaluate generated body gestures using FGD \cite{yoon2020speech} to measure distributional alignment with GT, reflecting realism. DIV \cite{li2021audio2gestures} quantifies gesture variation via the average L1 distance across clips. BC \cite{li2021ai} assesses speech-motion synchrony. For facial expressions, we use MSE \cite{yang2023diffusestylegesture} to quantify positional differences and LVD \cite{yi2023generating} to measure discrepancies between GT and generated facial vertices.

\subsection{Qualitative Results}
\textbf{Qualitative Comparisons.} We encourage readers to watch our demo video for a clearer understanding of SemTalk’s qualitative performance. Our method achieves superior speech-motion alignment, generating more realistic, diverse, and semantically consistent gestures than the baselines. As shown in \cref{fig:compare}, LivelySpeaker, TalkSHOW, EMAGE, and DiffSHEG exhibit jitter—EMAGE mainly in the legs and shoulders, while TalkSHOW affects the entire body. LivelySpeaker and DiffSHEG, which focus primarily on the upper body, produce slow and inconsistent motions, especially at speech clip boundaries. DiffSHEG improves gesture diversity over EMAGE and TalkSHOW, though EMAGE maintains greater naturalness. SemTalk surpasses all baselines in both realism and diversity. Compared to SemTalk*, SemTalk generates more expressive gestures, emphasizing key phrases (e.g., raising hands for “dream job” or pointing for “that is why”). While SemTalk* ensures rhythmic consistency, it lacks semantic expressiveness. By integrating frame-level semantic emphasis, SemTalk aligns motion with both rhythm and semantics, demonstrating the effectiveness of \textit{rhythmic consistency learning} and \textit{semantic emphasis learning}.
\begin{figure}
    \centering
        \includegraphics[width=0.48\textwidth]{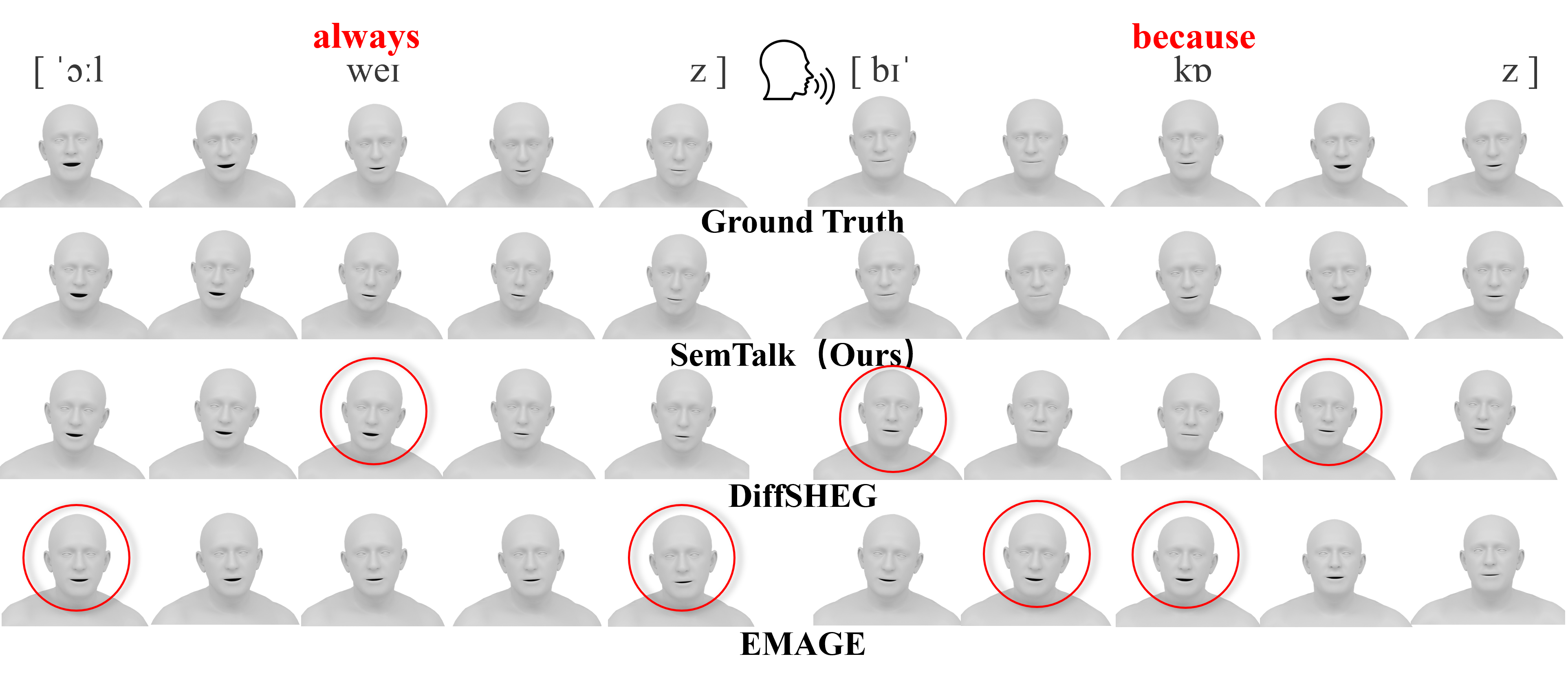}
        \caption{\textbf{Facial Comparison on the BEAT2 \cite{liu2024emage} Dataset.} 
        }
        \vspace{-15pt}
        \label{fig:face_compare}
\end{figure}
In facial comparisons (\cref{fig:face_compare}), EMAGE shows minimal lip movement, while both DiffSHEG and EMAGE reveal inconsistencies between lip motion and the rhythm of speech. In contrast, SemTalk produces smooth, natural transitions across syllables, resulting in realistic and expressive lips, significantly surpassing the baselines.

\begin{figure}
    \centering
        \includegraphics[width=0.43\textwidth]{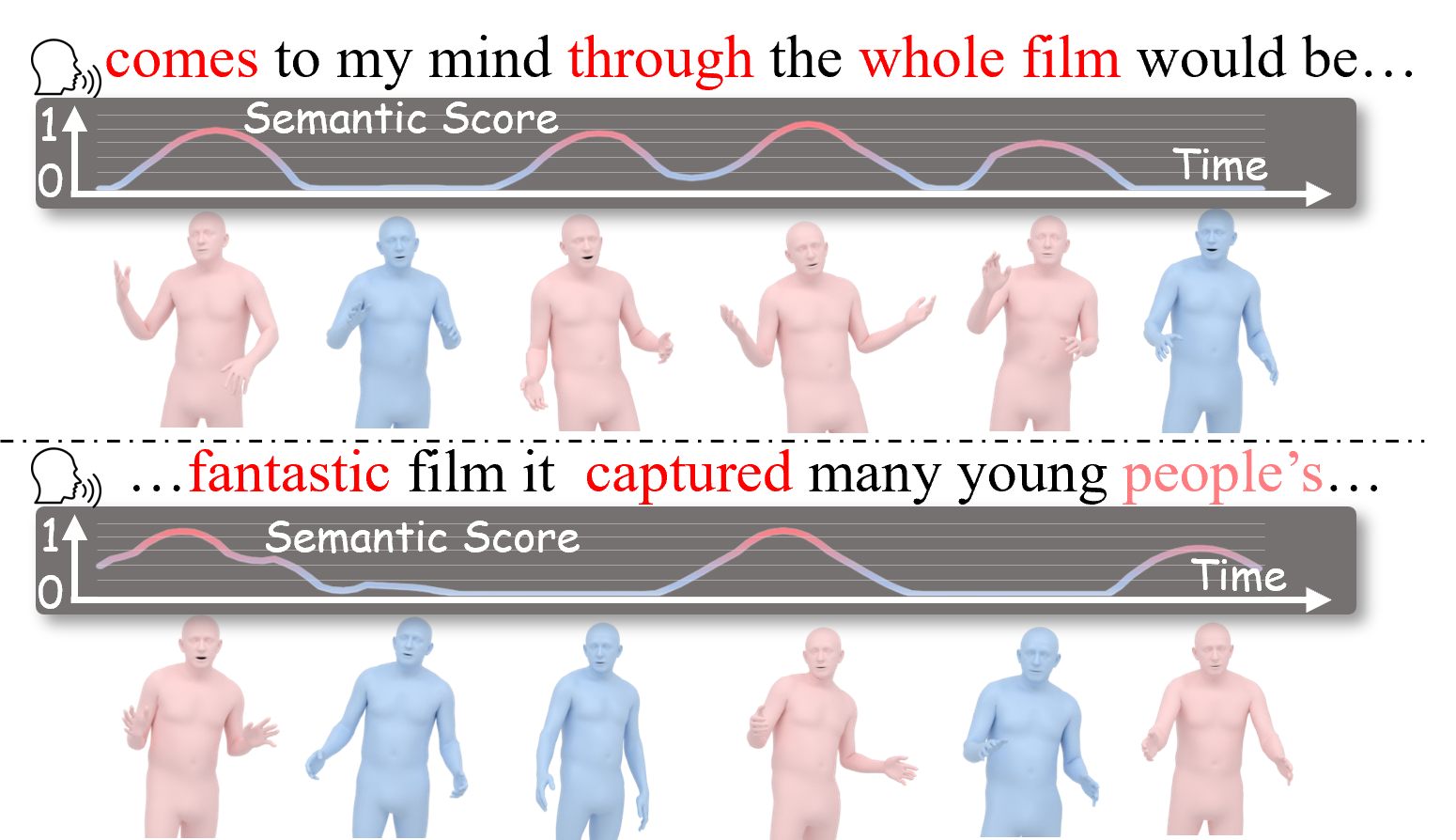}
        \caption{\textbf{Qualitative study on semantic score.} Semantic score aligns with keywords, influencing gesture intensity.}
        \vspace{-15pt}
        \label{fig:sem_score}
\end{figure}

On the SHOW dataset (\cref{fig:show_compare}), SemTalk shows more agile gestures than all baselines, when applied to unseen data. Our method captures natural and contextually rich gestures, particularly in moments of emphasis such as “I like to do” and “relaxing,” where our model produces lively hand and body movements that align with the speech content. 

\begin{figure}
    \centering
        \includegraphics[width=0.48\textwidth]{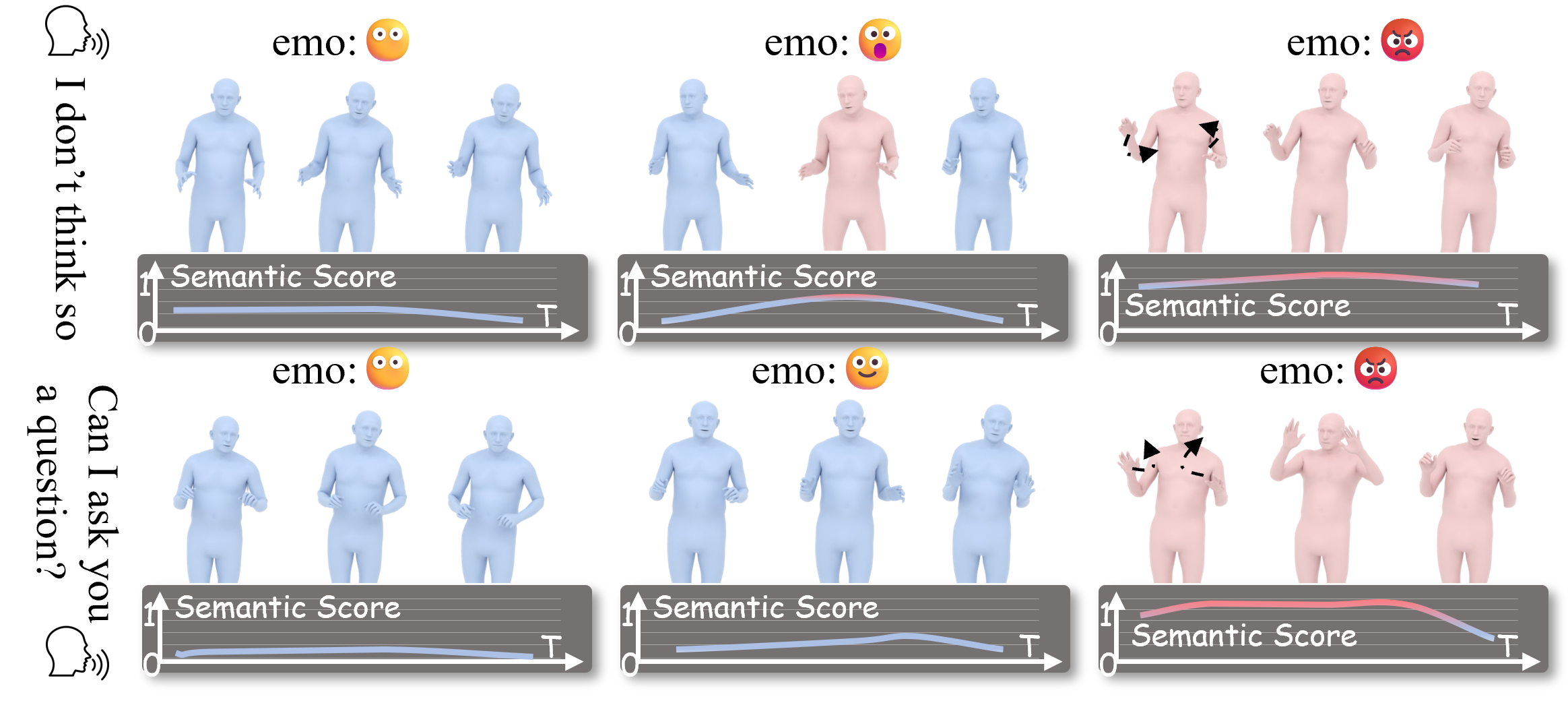}
        \caption{\textbf{Same words with different speech from the internet.} ``emo'' represents different emotional tones extracted from speech. SemTalk can generate different motions, even when the text script is the same, preventing overfitting to the text itself.}
        \vspace{-15pt}
        \label{fig:same_line}
\end{figure}

\noindent \textbf{Semantic Score.} \cref{fig:sem_score} shows how semantic emphasis influences gesture intensity, with peaks in the semantic score aligning with keywords like "comes," "fantastic," and "captured." By extracting semantic scores from key frames, we track gesture emphasis trends. Furthermore, as shown in \cref{fig:same_line}, SemTalk adapts to different emotional tones even when the text remains unchanged. This adaptability prevents overfitting to the text itself, allowing the model to generate gestures that vary according to the emotional delivery of the speech. The learned semantic score provides fine-grained, frame-level control, keeping gestures both rhythmically synchronized and semantically aligned in real time.

\begin{figure}
        \centering
        \includegraphics[width=0.43\textwidth]{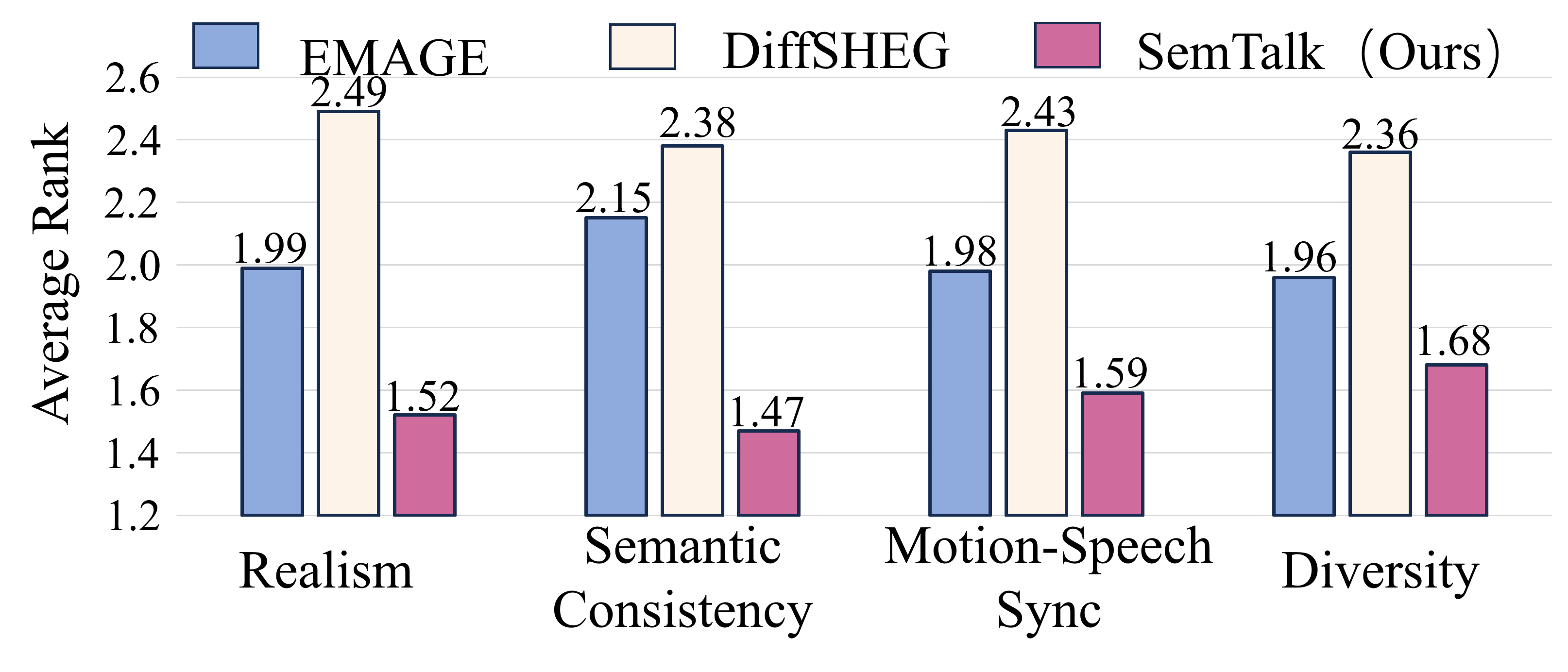}
        \caption{\textbf{Results of the user study.} }
        \vspace{-20pt}
        \label{fig:user_study_beat2}
\end{figure}

\noindent \textbf{User Study.} We conducted a user study with 10 video samples and 25 participants from diverse backgrounds, evaluating realism, semantic consistency, motion-speech synchrony, and diversity.  Participants were required to rank shuffled videos across different methods. As shown in \cref{fig:user_study_beat2}, our approach received dominant preferences across all metrics, especially in semantic consistency and realism. 
\subsection{Quantitative Results}
\noindent \textbf{Comparison with Baselines.}
As shown in \cref{tab:compare1}, SemTalk outperforms previous methods on BEAT2, achieving lower FGD, MSE, and LVD, indicating better distribution alignment and reduced motion errors. For fairness, we follow \cite{liu2024emage} and add a lower-body VQ-VAE to TalkSHOW, DiffSHEG, and SemTalk. Notably, SemTalk significantly reduces FGD, ensuring strong distribution matching. While TalkSHOW and EMAGE achieve competitive diversity (DIV) scores, SemTalk balances high semantic relevance with natural motion flow.

On the SHOW dataset, SemTalk excels with the lowest FGD, MSE, and the highest BC, indicating precise beat alignment with the audio and enhanced semantic consistency in generated motions. Although EMAGE exhibits high DIV, our model achieves comparable results while maintaining smooth, realistic motion free from jitter.

\begin{table}[t] 
   \centering
    \resizebox{0.46\textwidth}{!}{\begin{tabular}{clccccc}
        \toprule
        \textbf{Dataset} & \textbf{Method} & \textbf{FGD$\downarrow$} & \textbf{BC$\uparrow$} & \textbf{DIV$\uparrow$} & \textbf{MSE$\downarrow$} & \textbf{LVD$\downarrow$} \\
        \midrule
        \multirow{10}{*}{\rotatebox[origin=c]{90}{BEAT2}} & FaceFormer~\cite{fan2022faceformer} & - & - & - & 7.787 & 7.593 \\
        & CodeTalker~\cite{xing2023codetalker} & - & - & - & 8.026 & 7.766 \\
        & CaMN~\cite{liu2022beat} & 6.644 & 6.769 & 10.86 & - & - \\
        & DSG~\cite{yang2023diffusestylegesture} & 8.811 & 7.241 & 11.49 & - & - \\
        & LivelySpeaker \textit{et al.}~\cite{habibie2021learning} & 11.80 & 6.659 & 11.28 & - & - \\
        & Habibie \textit{et al.}~\cite{habibie2021learning} & 9.040 & 7.716 & 8.213 & 8.614 & 8.043 \\
        & TalkSHOW~\cite{yi2023generating} & 6.209 & 6.947 & \textbf{13.47} & 7.791 & 7.771 
        \\
        & EMAGE~\cite{liu2024emage} & 5.512 & 7.724 & 13.06 & 7.680 & 7.556 \\
        & DiffSHEG~\cite{chen2024diffsheg} & 8.986 & 7.142 & 11.91 & 7.665 & 8.673 
        \\
        \cmidrule(lr){2-7}
        & \textbf{SemTalk (Ours)} & \textbf{4.278} & \textbf{7.770} & 12.91 & \textbf{6.153} & \textbf{6.938} \\
        \midrule
        \multirow{10}{*}{\rotatebox{90}{SHOW}} 
        & FaceFormer~\cite{fan2022faceformer} & - & - & - & 138.1 & 43.69 \\
        & CodeTalker~\cite{xing2023codetalker} & - & - & - & 140.7 & 45.84 \\
        & CaMN~\cite{liu2022beat} & 22.12 & 7.712 & 10.37 & - & - \\
        & DSG~\cite{yang2023diffusestylegesture} & 24.84 & 8.027 & 10.23 & - & - \\
        & LivelySpeaker~\cite{habibie2021learning} & 32.17 & 7.844 & 10.14 & - & - \\
        & Habibie \textit{et al.}~\cite{habibie2021learning} & 27.22 & 8.209 & 8.541 & 145.6 & 47.35 \\
        & TalkSHOW~\cite{yi2023generating} & 24.43 & 8.249 & 10.98 & 139.6 & 45.17 \\
        & EMAGE~\cite{liu2024emage} & 22.12 & 8.280 & \textbf{12.46} & 136.1 & 42.44 \\
        & DiffSHEG~\cite{chen2024diffsheg} & 24.87 & 8.061 & 10.79 & 139.0 & 45.77
        \\
        \cmidrule(lr){2-7}
        & \textbf{SemTalk (Ours)} & \textbf{20.18} & \textbf{8.304} & 11.36 & \textbf{134.1} & \textbf{39.15} \\
        \bottomrule
    \end{tabular}}
    \caption{\textbf{Quantitative comparison with SOTA.} SemTalk consistently outperforms baselines across both the BEAT2 and SHOW datasets. Lower values are better for FMD, FGD, MSE, and LVD. Higher values are better for BC and DIV. We report FGD\(\times10^-1\), BC\(\times10^-1\), MSE\(\times10^-8\) and LVD\(\times10^-5\) for simplify.}
    \vspace{-10pt}
    
    \label{tab:compare1}
\end{table}
\noindent \textbf{Sem-gate.} ~\cref{tab:sem-gate} highlights the effectiveness of sem-gate. Without sem-gate, the model fails to emphasize key moments. Randomized semantic scores led to poor performance by preventing meaningful frame distinction.  Introducing a learned sem-gate even (w/o \(\mathcal{W}\)) significantly improves semantic alignment and classification accuracy. Refinement is further enhanced through weighting strategies: feature weighting \(\mathcal{W}_{f}\) enhances motion emphasis, while loss weighting \(\mathcal{W}_{l}\) improves FGD and overall accuracy. These results suggest that weighting methods enhance the accuracy of the semantic score and help the model prioritize important frames.  The best results come from applying two weighting methods together, where frames with stronger semantic signals receive higher emphasis.  We also compare sem-gate with LivelySpeaker's SAG \cite{zhi2023livelyspeaker}. We find that replacing the Sparse Motion stage with SAG and substituting motion using GT semantic labels led to poor performance. SAG relies only on text-motion alignment, ignoring emotional tone, making it more prone to overfitting the text. In contrast, our sem-gate applies GT supervision with two weighting methods, achieving more accurate and stable semantic motion. 

\noindent \textbf{Ablation Study on Components.}
We assess the impact of each component of our model on BEAT2 and present the results in \cref{tab:ablation}, which reveals several key insights (more ablation results please see supplementary material) :
\begin{itemize}
\item \textbf{Rhythmic Consistency Learning (RC)} not only boosts performance on key metrics like FGD, LVD, and BC but also reduces the MSE, contributing to smoother and more realistic base motion.

\item \textbf{Semantic Emphasis Learning (SE)} proves essential for selectively enhancing semantic-rich gestures. The inclusion of SE, as shown in rows with SE enabled, improves both diversity (DIV) and FGD, enabling the model to emphasize semantically relevant motions. SE demonstrates its effectiveness in focusing on frame-level semantic information, which contributes to the generation of lifelike gestures with enriched contextual meaning.
\begin{table}[t] 
   \centering
    \resizebox{0.46\textwidth}{!}{
    \begin{tabular}{ccccc}
        \toprule
         \textbf{Method} & \textbf{FGD$\downarrow$} & \textbf{BC$\uparrow$} & \textbf{DIV$\uparrow$} & \textbf{Acc(\%)$\uparrow$} \\
        \midrule
        w/o Sem-gate & 4.893 & 7.702 & 12.42 & -  \\
        SAG (LivelySpeaker \cite{zhi2023livelyspeaker}) & 4.618 & 7.682 & 12.45 & -\\
        Sem-gate (Random \(\psi\)) & 4.634 & 7.700 & 12.44 & 50.07  \\
        Sem-gate (w/o \(\mathcal{W}\)) & 4.495 & 7.633 & 12.26 & 72.32\\
        Sem-gate (w/ \(\mathcal{W}_{f}\)) & 4.408 & 7.679 & 12.28 & 78.52\\
        Sem-gate (w/ \(\mathcal{W}_{l}\)) & 4.366 & \textbf{7.772} & 11.94 & 77.83\\
        Sem-gate (ours) & \textbf{4.278} & 7.770 & \textbf{12.91} & \textbf{82.76} \\
        \bottomrule
    \end{tabular}
    }
    \caption{ \textbf{Ablation study on Sem-gate.} ``Acc'' denotes semantic classification performance on BEAT2. ``w/o Sem-gate'' means directly input \(f_{t}\) and \(\gamma_{h}\) without Sem-gate. ``SAG (LivelySpeaker \cite{zhi2023livelyspeaker})'' replaces the Sparse Motion Generation stage with LivelySpeaker's SAG method. ``Random \(\psi\)'' assigns frame-level scores randomly. ``w/o \(\mathcal{W}\)'' applies the semantic gate but excludes frame-level weighting. ``w/ \(\mathcal{W}_{f}\)'' applies feature weighting. ``w/ \(\mathcal{W}_{l}\)'' applies loss weighting. (as mentioned in \cref{sec:semantic-emphasis}). Sem-gate (ours) integrates both the semantic gate and frame-level weighting to enhance emphasis.}
\vspace{-15pt}
    \label{tab:sem-gate}
\end{table}
\begin{table}[t] 
   \centering
    \resizebox{0.46\textwidth}{!}{
    \begin{tabular}{ccccccccc}
        \toprule
         \textbf{RC} & \textbf{SE} & \textbf{C2F} &\textbf{RVQ} & \textbf{FGD$\downarrow$} & \textbf{BC$\uparrow$} & \textbf{DIV$\uparrow$} & \textbf{MSE$\downarrow$} & \textbf{LVD$\downarrow$} \\
        \midrule
        - & - & - & - & 6.234 & 7.628 & 11.44 & 8.239 & 7.831 \\
        - & - & - & $\surd$ & 5.484 & 7.641 & 11.84 & 13.882 & 15.42 \\
        $\surd$ & - & - & $\surd$ & 4.867 & 7.701 & 12.38 & 6.201 & 6.928 \\
        $\surd$ & $\surd$ & - & $\surd$ & 4.526 & 7.751 & 12.83  & 6.215 & 6.997 \\
        - & - & $\surd$ & $\surd$ & 4.897 & 7.702 & 12.42 & 13.416 & 15.72 \\
        $\surd$ & $\surd$ & - & - & 5.831 & 7.758 & 11.97 & 6.587 & 7.106 \\
        $\surd$ & - & $\surd$ & $\surd$ &4.397 & \textbf{7.776} & 12.49 & \textbf{6.100} & \textbf{6.898} \\
        $\surd$ & $\surd$ & $\surd$ & $\surd$ & \textbf{4.278} & 7.770 & \textbf{12.91} & 6.153 & 6.938 \\
        \bottomrule
    \end{tabular}
    }
    \caption{ \textbf{Ablation study on each key component.}
 “RC” denotes \textit{rhythmic consistency learning}, “SE” denotes the \textit{semantic emphasis learning}, and “C2F” denotes \textit{Coarse2Fine Cross-Att Module}, “RVQ" denotes the RVQ-VAE.}
\vspace{-10pt}
    \label{tab:ablation}
\end{table}
\item \textbf{Coarse2Fine Cross-Attention Module (C2F)} effectively refines motion details, improving BC, FGD, and DIV. When combined with RVQ and RC, C2F achieves the best MSE and LVD, highlighting its role in enhancing motion realism and diversity hierarchically.

\item \textbf{RVQ-VAE (RVQ)} enhances the diversity and realism of generated motion. Though it slightly increases MSE and LVD, it notably improves FGD, leading to more natural motion generation compared to standard VQ-VAE. 

\end{itemize}

\section{Conclusion}
We propose SemTalk, a novel approach for holistic co-speech motion generation with frame-level semantic emphasis. Our method addresses the integration of sparse yet expressive motion into foundational rhythm-related motion, which has received less attention in previous works. We develop a framework that separately learns rhythm-related base motion through \textit{coarse2fine cross-attention module} and \textit{rhythmic consistency learning}, while capturing semantic-aware motion through \textit{Semantic Emphasis Learning}. These components are then adaptively fused based on a learned semantic score. Our approach has demonstrated state-of-the-art performance on two public datasets quantitatively and qualitatively. The qualitative results and user study show that our method can generate high-quality co-speech motion sequences that enhance frame-level semantics over robust base motions, reflecting the full spectrum of human expressiveness.
{
    \small

}


\end{document}